\documentclass{vgtc}                          % final (conference style)
\ifpdf%                                % if we use pdflatex
  \pdfoutput=1\relax                   % create PDFs from pdfLaTeX
  \usepackage{graphicx}                % allow us to embed graphics files
  \DeclareGraphicsExtensions{.pdf,.png,.jpg,.jpeg} % for pdflatex we expect .pdf, .png, or .jpg files
\else%                                 % else we use pure latex
  \usepackage{graphicx}                % allow us to embed graphics files
  \DeclareGraphicsExtensions{.eps}     % for pure latex we expect eps files
\fi%

\graphicspath{{figures/}{pictures/}{images/}{./}} % where to search for the images

\usepackage{microtype}                 % use micro-typography (slightly more compact, better to read)
\PassOptionsToPackage{warn}{textcomp}  % to address font issues with \textrightarrow
\usepackage{textcomp}                  % use better special symbols
\usepackage{mathptmx}                  % use matching math font
\usepackage{times}                     % we use Times as the main font
\usepackage{cite}                      % needed to automatically sort the references
\usepackage{tabu}                      % only used for the table example
\usepackage{booktabs}                  % only used for the table example
\usepackage{amsmath}
\usepackage{amsfonts}
\DeclareMathAlphabet{\mathcal}{OMS}{cmsy}{m}{n}
\usepackage{tikz}
\usepackage{subfig}
\usepackage{multirow}
\usepackage{fancyhdr}
\usepackage{pgfplots} % package used to implement the plot  
\pgfplotsset{width=6.5cm, compat=1.6}  
\usetikzlibrary{patterns}

\newcommand{\etal}{\textit{et al}. }

\newcommand{\eg}{\textit{e}.\textit{g}. }

\onlineid{9332}

\vgtccategory{Application paper}
\vgtcinsertpkg

\title{Mixed Reality Communication for Medical Procedures: Teaching the Placement of a Central Venous Catheter}

\author{Manuel Rebol\thanks{e-mail: mrebol@american.edu}\\ %
    \parbox{1.4in}{\scriptsize \centering American University,\\Graz University of Technology}
\and Krzysztof Pietroszek\thanks{e-mail: pietrosz@american.edu }\\ %
     \scriptsize American University 
\and Claudia Ranniger\thanks{e-mail: ranniger@gwu.edu}\\ %
     \scriptsize George Washington University 
\and Colton Hood\thanks{e-mail: chood@mfa.gwu.edu}\\ %
     \scriptsize George Washington University %
\and Adam Rutenberg\thanks{e-mail: arutenberg@mfa.gwu.edu}\\ %
     \scriptsize George Washington University 
\and Neal Sikka\thanks{e-mail: nsikka@mfa.gwu.edu}\\ %
     \scriptsize George Washington University 
\and David Li\thanks{e-mail: dli@mfa.gwu.edu}\\ %
     \scriptsize George Washington University
\and Christian Gütl\thanks{e-mail: c.guetl@tugraz.at}\\ %
     \scriptsize Graz University of Technology 
     }

\teaser{
  \centering
  \includegraphics[width=\textwidth]{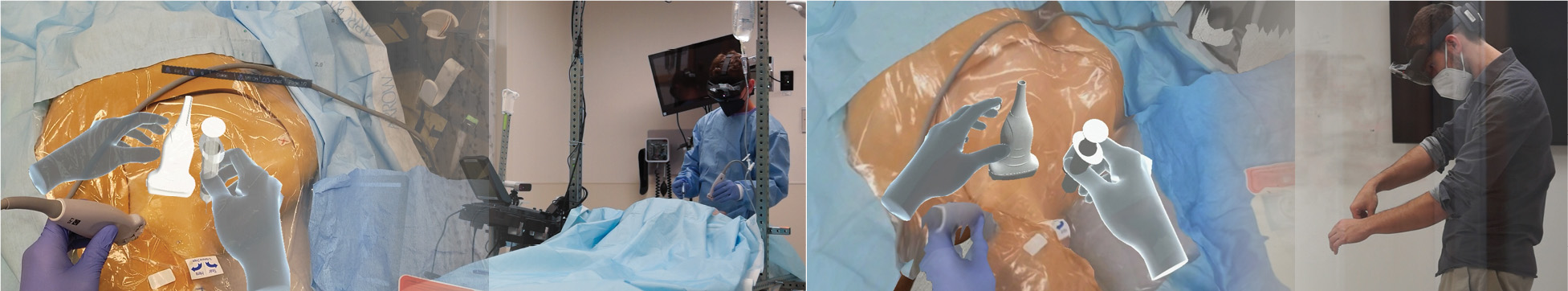}
  \caption{Mixed reality communication for medical procedures. We present a mixed reality communication system. A remote expert (right) guides a local operator (left) through the placement of a central venous catheter using augmented objects. 
  }
  \label{fig:teaser}
}

\abstract{%
Medical procedures are an essential part of healthcare delivery, and the acquisition of procedural skills is a critical component of medical education. Unfortunately, procedural skill is not evenly distributed among medical providers. Skills may vary within departments or institutions, and across geographic regions, depending on the provider’s training and ongoing experience.
We present a mixed reality real-time communication system to increase access to procedural skill training and to improve remote emergency assistance. Our system allows a remote expert to guide a local operator through a medical procedure. RGBD cameras capture a volumetric view of the local scene including the patient, the operator, and the medical equipment. The volumetric capture is augmented onto the remote expert's view to allow the expert to spatially guide the local operator using visual and verbal instructions.  
We evaluated our mixed reality communication system in a study in which experts teach the ultrasound-guided placement of a central venous catheter (CVC) to students in a simulation setting. The study compares state-of-the-art video communication against our system. The results indicate that our system enhances and offers new possibilities for visual communication compared to video teleconference-based training. } % end of abstract

\CCScatlist{
  \CCScatTwelve{Computing methodologies}{Computer graphics}{Graphics systems and interfaces}{Mixed / augmented reality};
  \CCScatTwelve{Computing methodologies}{Computer vision}{Computer vision problems}{Reconstruction}
  \CCScatTwelve{Social and professional topics}{Medical information policy}{Medical technologies}{Remote medicine};
  Telehealth; Volumetric communication;
}

\fancypagestyle{firststyle}
{
\fancyhead{}

\fancyfoot[CE,CO]{\footnotesize M. Rebol et al., "Mixed Reality Communication for Medical Procedures: Teaching the Placement of a Central Venous Catheter,"\\2022 IEEE International Symposium on Mixed and Augmented Reality (ISMAR), Singapore, Singapore, 2022, pp. 346-354, doi: 10.1109/ISMAR55827.2022.00050.}
\fancyfoot[LE,RO]{\thepage}
}

\begin{document}

\firstsection{Introduction} % intro + related studies: 1.5 pages

\maketitle
\thispagestyle{firststyle}
Patient care and procedural skills together form one of the Accreditation Council for Graduate Medical Education’s (ACGME’s) six core competencies for a practicing physician \cite{holmboe2016milestones}. Learning procedural skills typically requires that a trainee and an experienced medical professional are co-located. Training must be repeated periodically if the trainee does not perform the procedure regularly during training. Furthermore, the skills required to perform the procedure should be practiced periodically after training to avoid degradation \cite{https://doi.org/10.1002/aet2.10536}, especially if the operator does not perform the procedure regularly in his or her daily medical practice.
	
Multiple educational frameworks describe the acquisition of new procedural skills, but all have in common the iterative development of skill in a less experienced operator under the supervision and evaluation of a more experienced operator. Unfortunately, adequate procedural skill training is not always available to all medical providers. Thus, medical providers’ skills may vary depending on the department, institution, and geographic region. Nevertheless, some providers with limited experience may be placed in an emergency situation in which a procedure must be performed immediately \cite{telepresentintubation}. Consequently, there is an ongoing need to improve procedural skill acquisition and to provide remote assistance for medical providers with limited prior procedural experience or who work in resource-poor settings. Additionally, due to the geographical distribution of medical personnel, it is often difficult to arrange co-located training once the medical practitioner has completed initial medical training, especially for those who work in a remote areas, \eg in a critical access hospital. 

% An alternative is to provide distance training. Yet, few solutions have been proposed to date to resolve the gap between the quality and learning outcomes of distance training versus in-person training. The paucity of solutions presents a significant issue for equitable care delivery across health systems that have fewer experienced medical professionals.

% Based on the requirements, an elicitation study was performed. We hypothesize that the central problem in medical training at a distance may be related to the insufficient spatial information communication in the distance training environment. Pointing, gestures, demonstrations of actions and precise instructions, as well as positive and negative performance feedback, form a set of communication signals that are lost in the traditional, audio/video-based distance learning environment - yet we identify them as essential for effective training. 

\begin{figure*}[!ht]
    \centering
    \includegraphics[width=\textwidth]{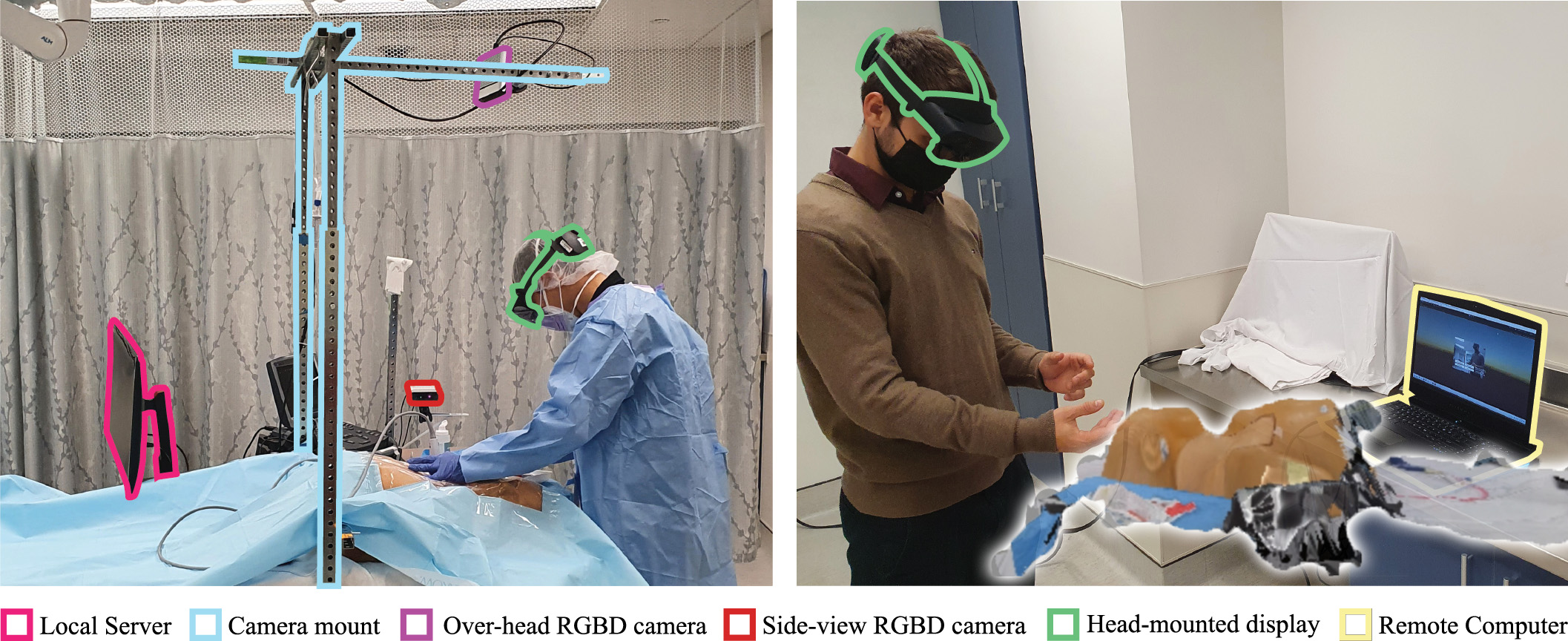}
    \caption{MR system overview for CVC placement. The local operator (left image) places a CVC while being assisted from a remote expert (right image) using the real-time mixed-reality communication system. We highlight the hardware components of the system.}
    \label{fig:sys-diagram}
\end{figure*} 
    
In order to address this issue, we will describe the design and implementation of a prototype real-time mixed-reality volumetric communication system that supports the acquisition of procedural skills for remote medical trainees. The system allows a remote expert to train and assist a medical trainee in learning a medical procedure without a need to be co-located with the trainee. We show examples of the different views of our communication system in \Cref{fig:teaser}.
We use the life-saving ultra-sound guided central venous catheter (US-CVC) placement procedure as an example procedure for our system design. We compare our mixed reality communication system against traditional video assistance in a user study. The participants complete the NASA Task Load Index (NASA-TLX) \cite{hart2006nasa} and open-ended surveys.

%%%%%%%%%%%%%%%%%%%%%%%%%%%%%%%%%%%%%%%%%%%%%%%%%%%%%%%%%%%%%%%%%%%%%%%%%%%%%%%%%%%%%%%%%%%%%%%%%%%%%%%%%%%%%%%%%%%%%%%%%%%

\section{Related Work} % intro + related studies: 1.5 pages

%related studies

Mixed reality (MR) poses a significant opportunity to enhance simulation-based training (SBT). Si et al. found that AR-based training simulations were able to accurately represent neurosurgical procedures, which is essential for a novice's comprehension and application of such training \cite{si2019assessing}. Shenai et. al. used the Virtual interactive presence and augmented reality (VIPAR) tool to provide telepresence virtual expert assistance during neurosurgical procedures. Using stereoscopic microscopes, both surgeon and expert were able to see both the surgical field and each other’s hands, with the remote expert able to provide visual and verbal guidance \cite{shenai2014virtual}. A simplified, non-stereoscopic system was subsequently used to support remote pediatric neurosurgical procedures with success \cite{davis2016virtual}. Similarly, Rojas-Muñoz et al. found that medical students were able to make incisions with greater accuracy using a telepresence AR system \cite{rohasmunoz2019surgical}.

A remote surgical assistance MR/AR system, known as ARTEMIS, was recently developed by Gasques et al. \cite{artemis}. This system provides a 3D representation of the expert to the student and is able to overcome many of the communication issues inherent with remote SBTs. For example, a remote expert is able to provide live annotations of the surgical field, while providing 3D hand gestures that can be visualized by the trainee and ultimately assist with the procedure. Initial evaluation of the system consisted primarily of qualitative feedback from study participants. The researchers found that novice trainees were able to successfully complete several complex surgical procedures while using the system. It is difficult, however, to assess how this system affected a trainee's cognitive load given the qualitative nature of the study. We propose a more affordable and less hardware-complex mixed reality communication solution compared to Gasques et al. Moreover, remote ultrasound (US) training adds additional challenges to the communication system \cite{Kessler2016_US2, article_US3,pietroszek2019univresity}. Mahmood \etal \cite{Mahmood2018_US1} proposed how US views can be used effectively in AR.

% Even fewer studies have used validated evaluation tools to assess procedures like a ultrasound guided central venous line insertion (US-CVC). Rochlen et. al. overlaid internal vascular anatomy on a mannequin to enhance central venous catheter insertion training \cite{rochlen2017first}. Independent raters used a checklist to grade learner knowledge and needle placement, however overall performance using a procedural checklist was not measured. They additionally used a survey instrument to elicit learner perceptions of the technology, which was positively received.
% Previous US-CVC studies have identified potential areas to augment the trainees SBT. Chen, et. al. demonstrated that during US-CVC placement, experienced operators fixate on the ultrasound image for significantly more time than novices, who focus on tools \cite{chen2018looks}. The ability of a mentor to independently view the procedural space is critical for patient safety. 

A first-person view of the procedural space has notable value in communicating elements of a trainee's environment to a remote expert. Some AR and MR systems have integrated this feature, which has been found useful by remote experts instructing trainees \cite{artemis}. Hand gestures themselves provide important non-verbal cues that may also add greater value to the trainee and remote expert interaction. Gestures may be used to direct trainees to a specific area of interest within the procedural space, or how to manipulate tools relevant to the procedure. Few studies have aimed to categorize such gestures in the context of surgical maneuvers for SBTs that involve MR or AR systems. Gesture recognition may increase the complexity of such a system but has utility in remote collaborative environments \cite{gesture}. Complex gestures, such as those found in medical procedure training, may be further analyzed into subcategories to add context during a remote collaboration. 

Ultimately, the preceding studies illustrate some current topics in the SBT MR and AR literature. Despite these novel findings, it is uncommon for studies to provide a synthesized and comprehensive solution that not only builds on the current state of AR/MR but also employs validated instrumental tools like the NASA-TLX. Also, studies rarely employ standardized gesture analysis within the context of medical SBT MR/AR systems. By employing validated tools, MR/AR systems may be rigorously tested for viability within medical SBT environments and reach the threshold for influencing patient care.

% related tech fundamentals:
To allow for volumetric communication, 3D scene reconstruction algorithms are used to combine multiple RGBD views to create a volumetric mesh. %Traditional mesh generation algorithms used tree structure depth map lookups. very slow. spatial point lookup: faster, real-time. 
Most recent algorithms utilize machine learning on large data sets \cite{gt-data-2017, data-benchmark2014} for high quality reconstruction of RGBD camera \cite{azinovic2021neural, OccupancyMappingand, optmization-network, Bozic_2020_CVPR} footage. Yet, high-quality reconstruction algorithms are too slow \cite{https://doi.org/10.1111/cgf.142651} for real-time communication. To overcome this issue, real-time 3D surface reconstruction has been proposed \cite{reconstruction, TextureMe, RealTimeGeometryAlbedo}. Chen \etal \cite{real-time-surface} achieve 24 Hz, for static scenes. For dynamic scenes, Yu \etal \cite{Yu_2021_CVPR} propose a real-time volumetric reconstruction algorithm to capture humans. Meertis \etal \cite{Real-time-scene-reconstruction-and2017} capture scenes with grid-based spatial and temporal depth map lookups for live volumetric view generation using desktop computers with high-end GPUs. %A similar approach was presented by Guo \etal \cite{RealTimeGeometryAlbedo}.
Meertis \etal filter the RGBD point cloud using a moving least squares (MLS) algorithm. Yet, the maximum time to create a mesh takes 167ms. We identified the need for low-latency and low-bandwidth visual communication that runs on mobile devices for medical training and emergencies.  Our proposed mesh generation algorithm has less than 1/30s latency and runs on mobile GPUs that are found in state-of-the-art head-mounted displays (HMDs). Moreover, our approach supports low-bandwidth connections.  

%%%%%%%%%%%%%%%%%%%%%%%%%%%%%%%%%%%%%%%%%%%%%%%%%%%%%%%%%%%%%%%%%%%%%%%%%%%%%%%%%%%%%%%%%%%%%%%%
%
\section{MR System Design} %2 pages

\begin{figure}
\centering
\resizebox{\columnwidth}{!}{%
 \begin{tikzpicture} 
        \node[draw={rgb,255:red,66; green,95; blue,237},rectangle, text width=1.7cm, align=center] at (0,3) (a) {Local\\HMD};
        \node[draw={rgb,255:red,66; green,95; blue,237},rectangle, text width=1.7cm, align=center] at (0,0) (b) {Local\\Server};
        \node[draw={rgb,255:red,191; green,31; blue,31},rectangle, text width=1.7cm, align=center] at (5,3) (c) {Remote\\HMD};
        \node[draw={rgb,255:red,191; green,31; blue,31},rectangle, text width=1.7cm, align=center,dashed] at (5,0) (d) {Remote\\Computer};
        \draw [semithick,->] (b) -- (a) node[midway,left, align=left, , xshift=28] {RGB US feed,\\Alignment (R,t)};
        \draw [semithick,->] (b) -- (c) node[midway,left, align=left, sloped, yshift=-5, xshift=19] {RGBD Volumetric,\\RGB US feed,\\Alignment (R,t)};
        \draw [semithick,->, dashed] (d) -- (a) node[midway,right, align=left, sloped, yshift=5, xshift=10] {RGB Video};
        \draw [semithick,->] (c) -- (a) node[midway, align=center] {Hand and Object\\Transformations (R, t)};
\end{tikzpicture}
}
\caption{Device and data exchange overview. We illustrate the four main communication devices using boxes. The arrows indicate the information transfer between the devices. Local and remote devices are highlighted in blue and red, respectively. The optional Remote Computer provides a video feed of the remote scene. 
}
\label{fig:network-diagram}
\end{figure}
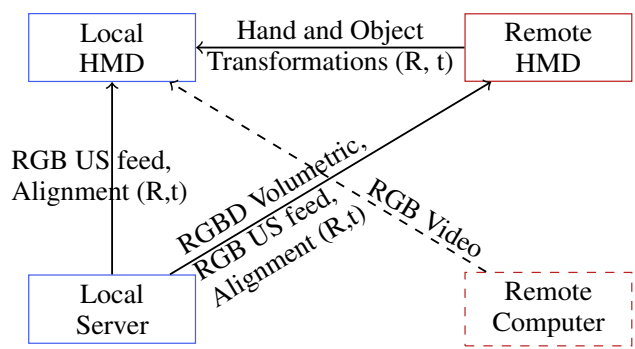

We propose a MR communication system for US-CVC training and emergency assistance to tackle the problem of unequal distribution of healthcare providers and procedural skills. We identified the initial design requirements for the mixed reality (MR) communication system in an elicitation study. In the elicitation study we analyzed in-person US-CVC training. We identified the need for spatial information, voice \& hands-free communication, aligned hand tracking, and virtual objects. Then, we started implementing and iteratively received feedback from medical experts to improve the system including the user interface and the augmented workspace setup.

During our system design phase, we tested different means of visual communication including a drawing and a pointing feature. However, we found virtual objects to be more useful for the US-CVC procedure. We also experimented with different alignment methods including markers and HTC Vive trackers. We found that our point correspondence method provides the best tradeoff between accuracy and setup time. %US CVC experts agreed that the alignment accuracy is sufficient for the procedure.
%For US-CVC assistance spatial information is essential. Hence, the proposed system needs to support volumetric capture. Moreover, sterility has to be maintained throughout the procedure, which suggests the use hands-free communication devices such as HMDs.    

% Integration
%Since during medical procedures, the cognitive load caused by the procedure itself is already high. Therefore, our objective is to design the system such that it does not increase cognitive load considerably. Our system does not require the user to have any specific knowledge about the technology. 

% The comm parties
We designed our system as a two-way real-time volumetric-based telepresence. It was designed to teach and support the US-CVC procedure, which is a procedure to place a large catheter into the central venous circulation. The procedure requires external landmark identification and psycho-motor skills to combine hand movements with information provided by ultrasound. Ultrasound is used to identify relevant anatomy and provides image guidance to facilitate a needle in puncturing the appropriate blood vessel which prevents injury. The two parties that our system is designed for are the remote expert and the local operator of the US-CVC procedure. The system supports a one-to-one connection between remote expert and the local operator. %For US-CVC training purposes, we propose a one-to-many connection such that a single expert is able to train multiple trainees at the same time.

\subsection{System Components}% How we address the requirements: System components
We address the requirements for mixed reality guidance of a US-CVC by designing the following components: 
\begin{itemize}
  \item We use a head-mounted augmented reality display for presenting the remote guidance to the local operator. The advantage of this technology is that information from the remote operator can directly be augmented on the local operator’s view. Thus, the local operator can focus on the procedure and does not need to use hands to operate the technology.
  \item Similarly, the remote instructor wears an AR head-mounted display (HMD) to be able to view a volumetric representation of the scene. Furthermore, the gestures and the interaction with the scene are
  captured by the HMD such that natural on-scene-like interaction is supported. 
  \item We deploy two volumetric cameras at the local site to present the local scene to the remote operator. We render the captured volumetric scene on the HMD the remote operator is wearing. 
  \item We use a microphone and speakers for voice communication between the remote expert and the local operator.
\end{itemize}
We illustrate the system including actors, devices, and communication flows in \Cref{fig:sys-diagram} and \Cref{fig:network-diagram}.

\begin{figure}
    \centering
    \includegraphics[width=\columnwidth]{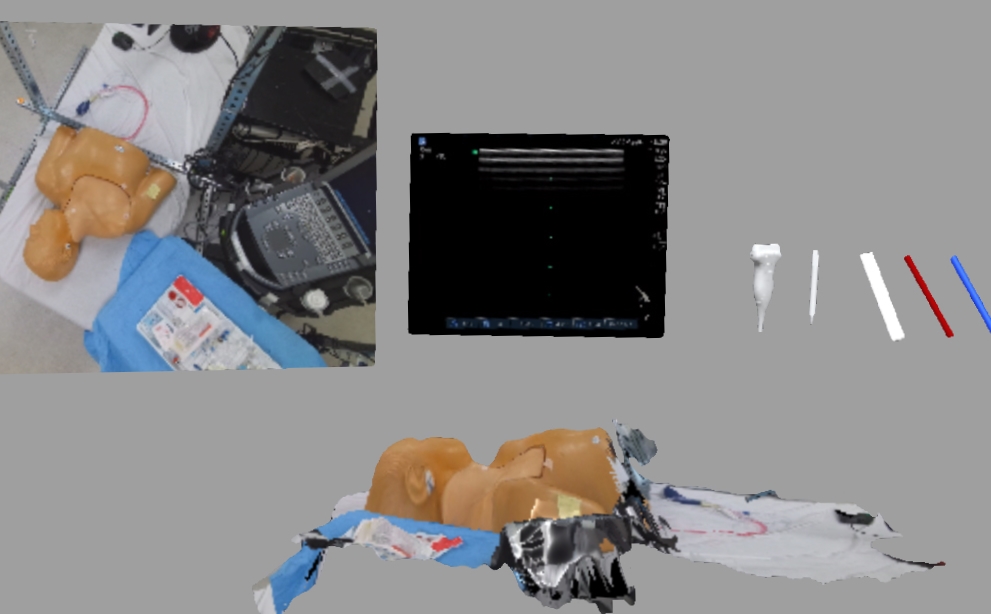}
    \caption{Remote view. The remote virtual workspace consists of the volumetric view (bottom center), the video feed (left), the ultrasound feed (top center), and the virtual objects (right). The virtual workspace can be placed in any room.}
    \label{fig:remote-view}
\end{figure}

\subsection{Devices} % The hardware
We use the Microsoft Hololens 2 HMD for augmenting the view of the local operator and for presenting the local view to the remote expert. In addition, the Hololens 2 captures the remote instructor's gestures that are sent to the local operator. The local scene is recorded by the volumetric camera Microsoft Azure Kinect. The procedure-specific ultrasound (US) feed is provided by a Sonosite M-Turbo machine. We deploy a small form factor computer on the local site to process the RGBD capture of the Azure Kinect camera and the US feed in real-time. Moreover, the computer acts as server that forwards the data between the Hololenses and the remote computer. %The remote computer renders the view for the remote operator's Hololens. In addition, the remote computer captures a video view of the remote scene. 

\subsection{Views} % Still high level, but more detailed explanation
The physical environment, as well as augmented information, are visible through the visor of the Hololens for the local operator and the remote expert. The augmented information allows for visual communication and interaction between the two actors.

\paragraph{Remote View}
The remotely located instructor receives the volumetric view recorded by cameras at the local site. It is augmented in 3D space on the instructor’s HMD. The volumetric view is the center of the augmented workspace. It is placed in the middle of the room the remote expert is situated in. The room needs to be empty because the remote expert needs space to interact with the volumetric view. We present the remote view in \Cref{fig:remote-view}.

Besides the volumetric view, the remote expert also sees a 2D video feed. The 2D video feed is captured by the same camera that also captures the volumetric view. Moreover, the remote expert sees the ultrasound feed from the local operators ultra-sound machine. 
The remote expert is able to manipulate a small set of virtual objects that can be used to instruct the local operator. These include abstract objects such as cuboids and cylinders as well as realistic renderings of medical tools used for the US-CVC procedure. These may be selected by the instructor, manipulated in space, resized, and shown to the local operator.

On top of the remote expert's hands, an augmented hand model is shown. This is the same hand model that is also shown to the local operator. It gives the remote expert a better understanding of how hand gestures look augmented on the local view. When the remote experts look at their palm, a hand menu is shown which gives them the options to switch between the cameras, enable and disable the virtual objects, and switch to long reach virtual arms.

\begin{figure}
    \centering
    \includegraphics[width=\columnwidth]{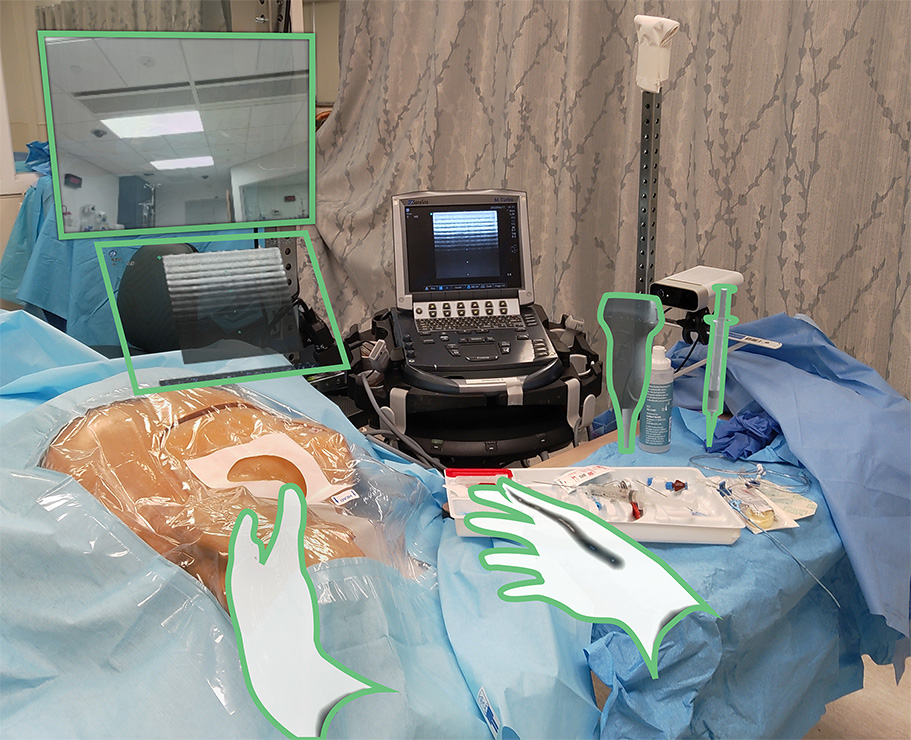}
    \caption{Local view. The physical workspace of the local operator includes the patient, the US device, and CVC-specific objects. The virtual view (highlighted using green borders for illustration) is used by the local operator to receive visual guidance from a remote expert.}
    \label{fig:local-view}
\end{figure}

\paragraph{Local View}
The physical environment plays an important role in the local view. The local operator needs to focus mostly on the physical workspace including the patient and medical instruments. Thus, the augmented visuals should only contain the most important information needed to get the required assistance from the remote expert. Moreover, we managed to eliminate all device interaction with the MR interface for the local operator. We present the local view in \Cref{fig:local-view}.

% What can the local operator see through the glasses 
The augmented information for the local operator includes two feeds in a static position. A video feed showing the remote expert and the ultrasound feed. Both feeds are located right above the physical area of the procedure. %The augmented ultrasound feed shows the same image as the physical ultrasound machine. However, the augmented feed is much closer to the procedural area. This allows the local operator to make smaller changes in focus when alternating attention between the ultrasound and physical area of operation while coordinating hand movements. 
The reason for the augmented US feed in our system is to provide the local operator with information close to the procedural area. This allows the local operator to make smaller changes in focus when alternating attention between the US and physical area of operation while coordinating hand movements. Alternatively, the local operator can use the physical US display. The position of the US feed was determined after consultation with US-CVC experts.
In addition to the static feeds, the local operator also sees the virtual objects and the instructors virtual hand \cite{Pietroszek2018b} model augmented by their HMD. The virtual objects can be moved by the remote expert. The virtual hand model moves as the remote operators hands move.  

%The local operator wears an AR HMD to receive visual guidance from the remote instructor. The gestures captured from the remote operator are visualized using a virtual 3D hand model which allows the local operator to receive directions.

\subsection{Interaction}
The visual interaction happens in both communication directions. The local operator's body language is captured through the Azure Kinect cameras and presented as a volumetric view to the remote expert. The remote expert has two options to guide the local operator: 
\begin{itemize}
    \item The expert can use hand gestures\cite{rebol2021passing, rebolGestures2021}, which are captured by the Hololens 2 camera system.
    \item The expert can use virtual objects, which can be manipulated by grabbing them with hands. 
\end{itemize}
The hand model can be used to give directions in the form of \eg pointing, showing how to hold instruments, and showing angles. The virtual objects can be used to show how to use them correctly including how to hold them and where to place them on the patient, and how to manipulate them in space. %%%%%%%%%%%%%%%%%%%%%%%%%%%%%%%%%%%%%%%%%%%%%%%%%%%%%%%%%%%%%%%%%%%%%%%%%%%%%%%%%%%%%%%%%%%%%

\section{MR System Implementation} % Implementation Details: 2 pages
% Network, View, Alignment
The three essential parts of our MR system are the network communication between the nodes, the 3D scene reconstruction and the alignment of the views between the two actors. For network connection, we use a peer-to-peer as well as a client-server architecture depending on the data communication types. The 3D scene captured by two Azure Kinect RGBD cameras is reconstructed using a grid mesh topology. The views are aligned between the remote and local operators using 3D point correspondences and head tracking to enable volumetric visual communication.   

\subsection{Network Setup and Data Communication Between Devices} % Don't mention local network aspect here. Only if needed, mention it at the Study setup section
From a networks perspective, we have four nodes in our system. We have the local computer, the local Hololens, the remote computer, and the remote Hololens. Each of the four nodes runs software we developed in Unity. The local computer also acts as server for initiating the peer-to-peer connections and establishing WebSocket connections. We show a diagram of the network and the data flow in \Cref{fig:network-diagram}. 

From a network perspective, we implement the Mixed Reality WebRTC client to manage the video data transfer from local server to remote Hololens, from local server to local Hololens and from remote computer to local Hololens. Moreover, we establish WebSocket connections between the local server, the local Hololens, and the remote Hololens. Through the WebSocket connection, we forward depth, transformation, and alignment data. We explain the data communication between the four devices below. 

\paragraph{Local Server}
%Communication/data aspects: high level
The local server provides three main services: network connection handling, local view capture, and local Hololens rending. As a network server, the local server initiates the WebRTC peer-to-peer connection and handles all WebSocket connections. We support unicast and multicast delivery of WebSocket data. The local view capture consists of the ultrasound feed and the volumetric view. Moreover, the local server renders the view for the local Hololens and sends it over network to the device. 

%Communication/data aspects: more detail
The volumetric view, consisting of color and depth frames are sent over the network through WebRTC and WebSocket channels, respectively. The frame number is encoded in both streams for synchronization purposes. We synchronize the color and the depth frames once received them from the Kinect camera and at the HMD of the remote operator. The mesh consistency is enforced before it is sent to the remote Hololens. The ultrasound input is transmitted as a video feed to the local and the remote Hololens. In addition to volumetric and US data, alignment information between the two Azure Kinect cameras is sent to the local and remote Hololens.

%device information
In terms of hardware, the local server is equipped with a state-of-the-art consumer CPU and GPU to allow for real-time processing.
Two Microsoft Azure Kinect RGBD cameras and the ultrasound machine are connected. The cameras capture color and the depth frames of the scene at 30 FPS. 

\paragraph{Remote HMD} 
The remote instructor receives color and depth frames from the local cameras and the US feed. The transformation between the two Azure Kinect cameras is sent to render the volumetric view at the remote site. Depth and color frames are received at 15 FPS. We found the following buffer and synchronization strategy to work best for our system. The remote HMD waits for up to 100 ms before rendering, if color and depth frames arrive out-of-order. After 100 ms, we skip the frame and move to the next frame. If the delay between the highest incoming depth and color frame number and the currently rendered frame number becomes more than 200 ms, we skip forward and continue with the highest frame number received to avoid latency from individual missing or out of order frames. The transformation information and the US feed are received at 15 FPS and 30 FPS, respectively.    

Our distributed 3D scene reconstruction algorithm (\ref{sec:3d-scene-reconstruction}) allows the mesh to be constructed from the incoming depth and color frames in real-time on the Hololens. Alternatively, the Hololens view can be rendered on the remote computer and sent with the Holographic Remoting Player \cite{holoremoting}. For interaction, the remote Hololens sends the transformation of hands and objects of the remote operator  %relative to camera 1 
to the local Hololens. 

\paragraph{Local HMD}
Initially, the local Hololens receives the object transformation information from the remote Hololens. Subsequently, it receives a video feed of the remote scene at 30 FPS over a peer-to-peer WebRTC connection. Finally, the US feed and the alignment information from the cameras are received from the local server. The incoming data is used to align and augment the remote information and the local US feed onto the local operator's view.

\paragraph{Remote Computer}
The remote computer captures the remote expert using a built-in webcam. The video feed is sent at 30 FPS via a WebRTC peer-to-peer connection to the local Hololens. If needed, the remote computer renders the view for the remote Hololens to decrease the load on the HMD. In case the remote video feed is not needed, the remote computer can be removed from the system.

\subsection{Distributed 3D Scene Reconstruction} \label{sec:3d-scene-reconstruction}
We implemented a distributed 3D dynamic scene reconstruction algorithm. Our lightweight algorithm runs on mobile GPUs and HMDs. The computation is distributed between the GPUs of the local server and the remote HMD. First, the temporal consistency is enforced by the local server. Second, the mesh is spatially smoothed on the remote HMD.

We explain the mesh generation process and our mesh optimization computations for each camera below. To merge the meshes generated by different cameras and to display the views on different physical locations, we apply the alignment computations presented in \Cref{sec:alignment}.

\paragraph{3D Scene Reconstruction per Camera}
We compute the volumetric representation for each RGBD camera individually. We display the volumetric representation of the local site on a grid of vertices $\pmb{V} \in \mathbb{R}^{288\times 320 \times 3}$. Each vertex on the grid can also be interpreted as a real-world point $P \in \mathbb{R}^3$. For each point $P \in \pmb{V}$ we read the corresponding depth and color value. We use multiple-view geometry \cite{hartley2003multiple} to read the correct depth and color values for each grid vertex. Intrinsic and extrinsic camera parameters are provided by the camera software. We provide a detailed description using a similar notation as in the OpenCV documentation \cite{opencv-camera} in the appendix.

As a result of the color and depth image pixel lookup, we get the 3D position of each vertex on our grid in meters relative to the depth sensor and its color value. 
We connect neighboring vertices on our grid to create a mesh. However, we only connect vertices less than 0.1 m away from each other. We linearly interpolate the color information from the image onto the mesh.

\begin{figure}%[h]
    \centering
    \captionsetup[subfigure]{labelformat=empty, position=top, font={small}}
    
    \subfloat{\raisebox{-1.5cm}{\rotatebox[origin=t]{90}{\small{t = 1}}}}\hspace*{3pt}%
    \subfloat[\small{0ms history}]{\includegraphics[width=0.315\columnwidth]{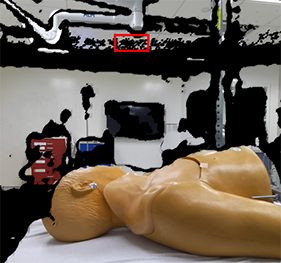} }%
    \subfloat[\small{200ms history}]{\includegraphics[width=0.315\columnwidth]{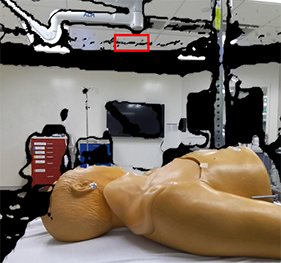} }%
    \subfloat[\small{1,000ms history}]{\includegraphics[width=0.315\columnwidth]{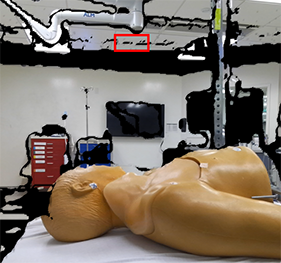}}%
    %\vspace{-0.3cm}
    
    % \subfloat{\raisebox{-1.5cm}{\rotatebox[origin=t]{90}{\small{200ms history}}}}\hspace*{3pt}%
    % \subfloat{\includegraphics[width=0.32\columnwidth]{fig/history/full-4-1.png}}%
    % \subfloat{\includegraphics[width=0.32\columnwidth]{fig/history/full-4-2.png}}%
    % \subfloat{\includegraphics[width=0.32\columnwidth]{fig/history/full-4-3.png}}
    % \vspace{-0.3cm}
    
    % \subfloat{\raisebox{-1.5cm}{\rotatebox[origin=t]{90}{\small{1000ms history}}}}\hspace*{3pt}%
    % \subfloat{\includegraphics[width=0.32\columnwidth]{fig/history/full-20-1.png}}%
    % \subfloat{\includegraphics[width=0.32\columnwidth]{fig/history/full-20-2.png}}%
    % \subfloat{\includegraphics[width=0.32\columnwidth]{fig/history/full-20-3.png}}
    
    %%%%%%%%%%%%%%%%%%%%%%%%%%%%%
    \vspace{0cm}
    \subfloat{\raisebox{-0.8cm}{\rotatebox[origin=t]{90}{\small{t = 1}}}}\hspace*{3pt}%
    \subfloat{\includegraphics[width=0.315\columnwidth]{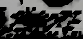} } %
    \subfloat{\includegraphics[width=0.315\columnwidth]{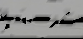} } %
    \subfloat{\includegraphics[width=0.315\columnwidth]{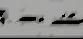}}
    \vspace{-0.3cm}
    
    \subfloat{\raisebox{-0.8cm}{\rotatebox[origin=t]{90}{\small{t = 2}}}}\hspace*{3pt}%
    \subfloat{\includegraphics[width=0.315\columnwidth]{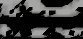} } %
    \subfloat{\includegraphics[width=0.315\columnwidth]{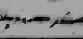} } %
    \subfloat{\includegraphics[width=0.315\columnwidth]{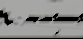}}
    \vspace{-0.3cm}
    
    \subfloat{\raisebox{-0.8cm}{\rotatebox[origin=t]{90}{\small{t = 3}}}}\hspace*{3pt}%
    \subfloat{\includegraphics[width=0.315\columnwidth]{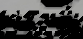} } %
    \subfloat{\includegraphics[width=0.315\columnwidth]{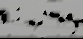} }%
    \subfloat{\includegraphics[width=0.315\columnwidth]{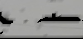}}
    \caption{Temporal consistency. We compare three different historic depth sensor reading settings (columns). The top row compares the temporal consistency on a full view. The bottom three rows compare a zoomed in view for three consecutive time steps. Note that increasing the historic reading time windows results in more scene information (top) and less noise (bottom).}
    \label{fig:history}
\end{figure} 

\paragraph{Grid Mesh Enhancements}  
The 3D grid mesh we created as explained above has a few visual deficiencies. They are due to the noise of the depth sensor and the nature of the grid topology. Because of the noisy depth data, the created grid mesh is unstable. Vertices move and some of them appear and disappear. Due to the nature of the grid topology, the edges are staircase-shaped. To improve the visual appearance of the 3D mesh, we used a two step process to enhance the mesh. First, we improve the stability of the mesh by temporal smoothing. This stability enhancement is computed on the local server, while maintaining the depth map structure before sending it to the remote Hololens for visualization. Second, we slightly correct the position of edge vertices on the mesh to create smoother object edges. This enhancement is computed on the remote computer that renders the mesh for the remote Hololens.  

The following Azure Kinect camera settings provide the best capture quality while keeping bandwidth requirements low for our system. We set the frame rate to 30 FPS and only keep synchronized color-depth image pairs. We set the color resolution to $1920 \times 1080$ and the depth resolution to $320 \times 288$ using the near field of view (NFOV) 2x2 binned (SW) depth model \cite{azure-hardware}.

%History on local server
We enforce temporal consistency on the local server. We replace invalid sensor readings by the latest historic reading of the last 200ms for each depth pixel at position $(i,j)$. We show an example of how different historic reading times lead to a more stable mesh in \Cref{fig:history}. We measure the effect quantitatively by capturing one second. 200ms of historic reading recovers 4\% of the lost vertices and reduces the on/off vertex flickering by 45\%. Increasing the time to 1000ms recovers 7\% of the lost vertices and reduces the on/off vertex flickering by 67\%. We found that 200ms works best for both static and dynamic scenes. 

We tackle small jitter by computing the moving average $\Bar{d}$ of the valid depth sensor readings $\mathcal{D}(\cdot)> 0$ over the last $n=10$ frames
\begin{equation}
   \Bar{d}(i,j) = (\sum^{-1}_{t=-n} \mathcal{D}(i,j,t)) \ / (\sum^{-1}_{t=-n} \delta (\mathcal{D}(i,j,t)> 0)),
\end{equation}
where $\delta(\cdot)$ refers to the indicator function:
\begin{equation} %\small 
\delta(\phi(\cdot)) = \begin{cases} 1 & \text{if } \phi(\cdot) \text{ is true } \\ 
0 & \text{else.} \end{cases} 
\end{equation}
In case the current depth value $\mathcal{D}(i,j,0)$ is within 3mm of the moving average $\Bar{d}$, we assign the previous depth value $\mathcal{D}(i,j,-1)$ to stabilize the vertex. We found $n=10$ and a 3mm moving average work best for the CVC procedure setup.
The small jitter stabilization reduced the mean per frame jitter from 128m to 67m (-48\%).

%POSS: write about additional optimization (_convergeNoisy). 

We tackle large jitter by counting the number of changes greater than $\lambda = 3$mm within the last 60 frames $n_2 = 60$,
\begin{align}
    \xi(i,j) = \sum^{-2}_{t=-n_2} &\delta(|\mathcal{D}(i,j,t) - \mathcal{D}(i,j,t+1)| > \lambda) \ \cdot \\
   &\delta(\mathcal{D}(i,j,t) > 0 ) \cdot \delta(\mathcal{D}(i,j,t+1) > 0 ) \notag.
\end{align}
Similar to small jitter, we assign the previous depth value $\mathcal{D}(i,j,-1)$ if $\xi(i,j) / n_2 > 0.6$. The 0.6 threshold in combination $\lambda = 3$mm and $n_2 = 60$ produced the best results empirically.

The three mesh enhancements explained above (historic reading, small jitter, and large jitter) result in more temporally stable vertices and lower bandwidth requirements. The enhancements combined reduced the mean per frame jitter from 128m to 40m (-68\%) in our CVC procedure setup.

\begin{figure}
    \centering
    %\subfloat[\small{Grid Mesh}]{\includegraphics[width=0.5\columnwidth]{fig/edges/edges-raw-rect-far.png}\label{subfig:default-full}}%
    %\subfloat[\small{Moved + Feathered Edge Mesh}]{\includegraphics[width=0.5\columnwidth]{fig/edges/edges-move-rect-far.png}\label{subfig:all-full}}
    \subfloat[\small{Grid Mesh}]{\includegraphics[width=0.5\columnwidth]{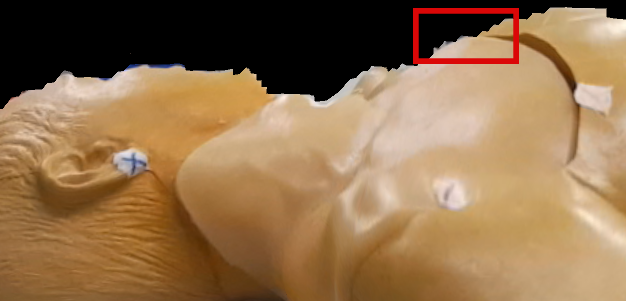}\label{subfig:default-full}}%
    \subfloat[\small{Moved + Feathered Edge Mesh}]{\includegraphics[width=0.5\columnwidth]{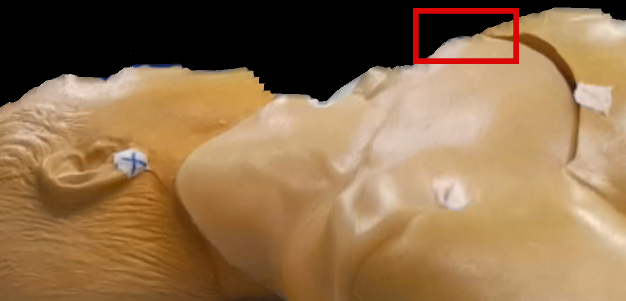}\label{subfig:all-full}}
    
    \subfloat[\small{Grid}]{\includegraphics[width=0.25\columnwidth]{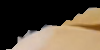}\label{subfig:default}}%
    \subfloat[\small{Feather}]{\includegraphics[width=0.25\columnwidth]{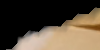}\label{subfig:feather}}%
    \subfloat[\small{Move}]{\includegraphics[width=0.25\columnwidth]{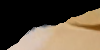}\label{subfig:move}}%
    \subfloat[\small{Feather + Move}]{\includegraphics[width=0.25\columnwidth]{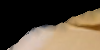}\label{subfig:all}}%
    \caption{Edge refinement. An example of edge refinement on a mesh side-view of the CVC mannequin. We compare the default grid mesh after combining color and depth information \protect\subref{subfig:default-full} with the refined mesh \protect\subref{subfig:all-full}. The individual refining steps are illustrated in detail on a zoomed-in view: grid mesh \protect\subref{subfig:default}, feathered edges \protect\subref{subfig:feather}, moved vertices \protect\subref{subfig:move}, and feathered edges combined with moved vertices \protect\subref{subfig:all}. 
    }
    \label{fig:edges}
\end{figure}

% Stuff on remote computer: move edges, alpha
After the depth data used to construct the mesh is received by the remote computer, we apply additional mesh enhancements on the vertex level to improve the edge appearance. We show an example of edge refinement in \Cref{fig:edges}.
We move the vertices to on-edge positions to remove unnatural edges created by the grid-aligned mesh. Therefore, we consider the 8-neighborhood of adjacent vertices for each vertex. Depending on the number of neighbor vertices within a 10 cm distance, the central vertex is moved in a direction that produces a natural edge. The movement takes into account the grid topology of the mesh. We show our grid topology and examples of how we modify the vertices depending on the number of neighbors in the appendix. %in \Cref{fig:nh}.

% \begin{figure}
%     \centering
%     \includegraphics[width=\columnwidth]{fig/nh.png}
%     \caption{Vertex manipulation. We consider each central vertex (orange dot) and 8 neighboring vertices (blue dot) numbered clock-wise starting at the top left. The first square (top right) shows the triangulation topology to create a mesh. The other squares show how we modify the central vertex position to refine the edges of the mesh taken into account the grid topology. Depending on how many neighbors $n$ a vertex has, a new vertex position (green dot) is assigned. We show two example cases A and B for the per neighbor count. Blue lines indicate original neighbor presence and green lines indicate the final mesh connections from the central vertex after it was moved. Note that this is a 2D representation of the 3D grid $V$.}
%     \label{fig:nh}
% \end{figure} 

In addition, we set the alpha value of edge vertices depending on their number of vertices. We set the alpha value of every vertex by dividing the number of neighbor vertices within the 10cm distance by 8. This feathering of object edges allows the edges to appear more natural on the remote Hololens. 

We only show triangles in the constructed mesh with an edge length smaller than 10cm. This allows us to remove inaccurate information about the scene from a view angle of the scene not captured by the RGBD cameras.

\subsection{Positioning and Alignment Between Actors} \label{sec:alignment}
%%% Continue here. 
%On the camera computer, we deploy the Microsoft Kinect SDK which imports the captured color and depth data into Unity 3D game engine. We use OpenCV for Unity for the Aruco marker detection and pose estimation. We implement a Kalman filter to remove the noise of the Aruco detection as shown by \cite{KalmanAruco}.
We align the physical and virtual workspace between the local instructor and the remote expert to enable volumetric communication. First, we focus on the alignment between the two 3D meshes created by the Azure Kinect cameras on the local site. Aligning them allows us to create a volumetric view. Second, we align the volumetric view with the physical world on the local site using point correspondences. The camera and the physical alignment together with the built-in head tracking of the Hololens 2 allows for volumetric communication using pointing, gestures, and virtual tools.    

\paragraph{3D View Alignment} 
%The local scene uses an Aruco marker to solve two common alignment problems. First, it allows the freedom for the remote operator to position the volumetric capture at the desired location. Second, the marker is used to align the capture of the volumetric camera with the current view of the local trainee.   
%POSS: Aruco vs HTC Vive vs Markers and Balls

% Camera 1 and camera 2 alignment; physical view and camera 1 alignment
For finding the rigid transformation $\pmb{R}$ and $t$ between the 3D meshes, we apply least-squares fitting \cite{leastSqaure} using $N=4$ point correspondences. Four correspondences result in high alignment accuracy (see \Cref{sec:system-evaluation}) while keeping the setup time low.   
Least-squares fitting minimizes the error
\begin{equation} \label{eq:ls}
    \epsilon = \sum_{n=1}^N || \pmb{R} \pmb{A}^n + t - \pmb{B}^n || ^ 2 ,
\end{equation}
where $\pmb{A} \in \mathbb{R}^{N \times 3}$ and $\pmb{B}\in \mathbb{R}^{N \times 3}$ are sets of 3D point correspondences from the 3D meshes. First, we compute the centroids of each point set using 
\begin{equation}
    \psi(\pmb{S}) = \frac{1}{N} \sum_{n=1}^N \pmb{S}^n .
\end{equation}
We subtract the centroid of each point set to center the points around the origin. Then, multiply the two centered point clouds and apply singular value decomposition $SVD(\cdot)$ to find the rotation matrix $\pmb{R}$:
\begin{align}
    (\pmb{U}, \pmb{S}, \pmb{V}) &= SVD( (\pmb{A}-\psi(\pmb{A}))(\pmb{B}-\psi(\pmb{B}))^T ) \\
    \pmb{R} &= \pmb{V}\pmb{U}^T \notag .
\end{align}
Once we checked for reflection $|\pmb{R}| < 0$, we compute the translation 
\begin{equation}
    t = \psi(\pmb{B}) - \pmb{R} \psi(\pmb{A}) .
\end{equation}
The resulting rotation $\pmb{R}$ and translation $t$ describe the rigid transformation between different camera or world views given 3D point correspondences $\pmb{A}$ and $\pmb{B}$. 

We use the transformation computation presented above for aligning the two camera views on the local site to create a volumetric representation at the remote site. Moreover, we compute the rigid transformation between the physical local Hololens device and camera 1 to align remote gestures and objects to the physical world of the local operator.

% CHECK FONT FAMILY
% \makeatletter
% \edef\textFontName{\fontname\csname
%   \f@encoding/\f@family/\f@series/\f@shape/\f@size\endcsname}
% \edef\mathFontName{\fontname\textfont0}
% \edef\mathLetterFontName{\fontname\textfont1}
% %\make

% See\footnote{Text Font: \textFontName

% Math Operator Font: \mathFontName

% Math Letter Font: \mathLetterFontName\par
% }

\paragraph{Gestural Communication}
We built our AR application for the Hololens 2 HMDs using the Mixed Reality Toolkit. The toolkit allows us to support standardized AR user interaction. On the remote instructor's Hololens, we used Mixed Reality Toolkit's pose detection to predict the hand position. The head-tracking is also used for alignment between the nodes. 

For gestural communication, we send a hand model representing the remote expert's hands to the local operator. We detect the hand gestures using the camera system of the Hololens 2. The 3D position and rotation of 26 hand joints are sent to the local operator. We send the hand joint data $\{R,t\}_1$ in camera 1 coordinates:
% \begin{align}
%     \pmb{R}_{\text{HandJointC1}} &= \pmb{R}^{-1}_{\text{RemoteVolViewTransform}} \pmb{R}_{\text{HandJointRemoteHololens}} , \\
%     t_{\text{HandJointC1}} &= \pmb{R}^{-1}_{\text{RemoteVolViewTransform}} (t_{\text{HandJointRemoteHololens}} \notag \\ 
%   & \qquad \qquad \qquad \qquad \qquad - t_{\text{RemoteVolViewTransform}}) . \notag
% \end{align}
\begin{align} \label{eq:transform}
    \pmb{R}_1 &= \pmb{R}^{-1}_2 \pmb{R}_3 , \\
    t_1 &= \pmb{R}^{-1}_2 (t_3 - t_2) , \notag
\end{align}
where $\{R,t\}_2$ refers to the remote volumetric view transform, and $\{R,t\}_3$ refers to the hand joints in remote Hololens coordinates.  
At the local site a hand model is animated to represent the remote instructor's gestures using \Cref{eq:transform} where $\{R,t\}_1$ represents the hand joints in local Hololens coordinates, $\{R,t\}_2$ represents the local to camera 1 transformation, and $\{R,t\}_3$ represents the hand joints in camera 1 coordinates.
% \begin{align}
%     \pmb{R}_{\text{HandJointLocalPhysical}} &= \pmb{R}^{-1}_{\text{LocalPhysicalToC1}} \pmb{R}_{\text{HandJointC1}} , \\
%     t_{\text{HandJointLocalPhysical}} &= \pmb{R}^{-1}_{\text{LocalPhysicalToC1}} (t_{\text{HandJointC1}} \notag \\ 
%   & \qquad \qquad \qquad  \qquad  - t_{\text{LocalPhysicalToC1}}) . \notag
% \end{align}
We estimate $\{R,t\}_2$ using the point correspondence optimization illustrated in \Cref{eq:ls}. 
We apply the same transformation for virtual objects sent from remote to local.
% CHI paper: detailed vs abstract models, objects

%TODO: long arm

%POSSible: add offset for MRTK model

%POSS: add correction rotations and stuff

\paragraph{Initial Setup Procedure}
We propose an initial calibration phase during system setup to align the RGBD camera coordinate systems and the Hololens coordinate systems. For convenience, 4 markers are stuck on a static object in the local scene, close to the area of interest. Note that any landmark can be used instead of markers for this step. Both RGBD cameras see the markers. The local operator then places virtual points on top of the markers for each camera on the local server AR application. The local Hololens camera system is also calibrated using the markers. The references between physical marker, Hololens, and Azure Kinect camera 1 enable volumetric collaboration. 

The virtual workspace setup on the local Hololens is relative to the physical markers as shown in \Cref{fig:local-view}. The remote expert initiates the workspace standing behind the desired workspace location. The virtual workspace appears in front of them similar to a monitor setup as shown in \Cref{fig:remote-view}. The remote expert then fine-tunes the position and rotation of the volumetric view using hand gestures.

\section{Evaluation} % Evaluation + Conclusion: 1.5pages
We evaluated the proposed mixed reality system in a study in which medical experts taught the ultrasound-guided placement of a central venous catheter to learners. We conducted the study in a simulation center using CAE Blue Phantom %\textsuperscript{\textregistered} 
ultrasound central venous access training mannequins \cite{mannequin}. %We measured the cognitive load during the procedure for instructors and learners. 
We compared training with the proposed MR system against training with video communication software. For video communication, we capture the same views as the MR system, a side, and an over-the-shoulder view. %We compared the traditional video training against training through our mixed reality communication system using
The results were analyzed using surveys and video recordings.

\subsection{System Setup Analysis} \label{sec:system-evaluation}
We constructed a modular camera mount to allow for flexible camera positioning. Yet, the camera mount is rigid and provides stable camera views throughout the procedure. Our camera mount is attached to the stretcher using existing bolts.  This allows for fast setup and consistent camera positions relative to the mannequin between sessions. Moreover, we added a pin mount to the base of the camera mount to secure the mannequin position on the stretcher. After we evaluated different views with domain experts, we found that a side combined with a top camera position provides the best view for the procedure. %We measure the position of the depth sensor relative to the mannequin's laryngeal prominence. %We positioned the side camera at 86cm away, 10cm to the top of the stretcher, 85cm to the right side of the patient and 5cm up. We positioned the top camera at 103cm distance. 25cm towards the top of the stretcher, 16cm to the right, and 99cm up. 
We mounted the side camera and the top camera at distances of 86cm and 103cm, respectively, relative to the laryngeal prominence of the mannequin. 
Both cameras were rotated such that they point to the area relevant for the US-CVC procedure.

%POSS: exact AR workspace setup remote and local: m size, pixels,...
We measured the alignment accuracy between the local and remote after the system setup. Each time, we took four measurements evenly distributed around the borders of the tissue insert from the CVC mannequin, at the critical area of the procedure. 
We found the mean error between the remote operator's volumetric view and the local physical to be 1.36cm, $\sigma = 0.18$cm, for $n = 10$ setups. The participants of our study reported that the accuracy is sufficient for giving pointing instructions and visual object guiding. The alignment error after the initial setup between the two camera views on the remote side is 2.54cm, $\sigma = 0.34$cm, for $n = 5$ setups. We lower the initial in-between camera error manually. This is possible because of our static camera setting on the local site.  
%Note that we are able to lower the error by manual correction after the automated ball setup procedure below 1cm because of static camera positioning on the mount. However, we found that the automated setup alignment accuracy is high enough for instructing CVC over MR. 

\subsection{Study Participants}
Our study participants consisted of a group of 5 instructors and 20 learners, all lived and trained in the USA. We randomly assigned the instructors and the learners to 10 mixed reality and 10 video training sessions. Each learner completed exactly one training session whereas instructors taught multiple sessions. Each instructor taught at least one video and one mixed reality session. We illustrate the learner demographics and prior experience in \Cref{tab:learner-demographics}.
The instructors were on average 43 years old and consisted of four males and one female. All of them were performing US-CVC for more than 3 years and teaching the procedure for more than 1 year. 

% \begin{table}[]
% 	\setlength\extrarowheight{2pt} 
% 	%\setlength{\tabcolsep}{2.6pt}
% 	%\scriptsize
% 	\footnotesize
% 	\centering
% 	\begin{tabular}{l|c|c}
% 		\textbf{Group} & \textbf{Video} &\textbf{MR}\\ \hline \hline
% 		{Age} & 26.5y & 27.4y\\ \hline
% 		{Male/Female} & 2/8 & 4/6\\ \hline
% 		{Clinical training and/or practice experience} & 1.8y & 2.2y\\ \hline
% 		{Prior AR/MR/VR experience} & 0 & 0\\ \hline
% 	\end{tabular}
% 	\caption{Learner demographics and prior experience.
% 	}\label{tab:learner-demographics}
% \end{table}	
 
\begin{table}
	\setlength\extrarowheight{2pt} 
	\setlength{\tabcolsep}{12pt}
	%\scriptsize
	\footnotesize
	\centering
	\begin{tabular}{l|c}
		%\textbf{Attribute} & \\ \hline \hline
		\hline \hline
		{Age} & 26.9y \\ \hline
		{Male/Female} & 6/14\\ \hline
		{Clinical training and/or practice experience} & 2y\\ \hline
		{Prior AR/MR/VR experience} & 0\\ \hline \hline
	\end{tabular}
	\caption{Learner demographics and prior experience.
	}\label{tab:learner-demographics}
	%\vspace{-6pt}% TODO: just to fit the 8pages
\end{table}

\subsection{Study Setup}
 We prepared a Blue Phantom ultrasound central venous access training mannequin \cite{mannequin}, a CVC kit, and a Sonosite M-Turbo Ultrasound system \cite{us}. 
  The following parts of the CVC procedure were taught:
  \begin{enumerate}
      \item A talk through the procedural steps, the preparation of the CVC kit, and the use of the ultrasound.
      %\item The .
      \item Catheter placement over a wire using the Seldinger technique facilitated through the catheter over the needle approach. Confirmation of wire placement is achieved with ultrasound. 
      \item The flushing and drawing of the three catheter ports after insertion.
  \end{enumerate}
 We compared training with the proposed MR system against training with video communication software. For video communication, we capture the same views as the MR system, a side, and an over-the-shoulder view. 
 %We performed ten training sessions for cognitive load evaluation for each methodology video and MR. We assigned 10 students and 5 doctors randomly to the individual sessions.  

\subsection{Study Process} 
Prior to the study, learners and instructors provided informed consent. The learners completed US-CVC pre-training to familiarize themselves with the steps of the procedure. Instructors and learners had not received MR training prior to the study. The learners completed a pre-training survey which included demographic and prior experience questions. At the beginning of each training session, instructors talked about background information on US-CVC with the learner. Then, they prepared the learner's workspace for the procedure and talked through the medical equipment necessary for the procedure. After the initial preparation, instructors moved to a separate room to start with the video or mixed reality training.  

In the case of MR sessions, both instructors and learners completed a 5-minute MR briefing. Apart from the briefing, the participants did not receive any training on the technology. Then, the actual training session, which took about one hour, started. After the US-CVC session, both learners and instructors completed surveys and interviews.  

\subsection{Study Results}
We analyzed post-session interviews, recorded video data, and NASA-TLX survey responses to evaluate the mixed reality system. Overall, the feedback from both instructors and learners was very positive on using the mixed reality communication system for US CVC training.  

\paragraph{Workload Analysis}
% The workload for learner and instructor was measured using a NASA Task Load Index (TLX) survey. We show the results in  %\Cref{tab:nasa-tlx} and 
% \Cref{fig:nasa-tlx}. We performed a two-sample two-tailed t-test assuming a normal distribution. Using a significance level of $\alpha = 0.05$, we could not find a significant difference in total as well as in per category workload between MR and video. 

The NASA-TLX gave us a quantitative subjective measure of the workload for instructors and learners. The validated instrument allowed us to compare video and MR sessions, see \Cref{fig:nasa-tlx}. Our null hypothesis was that video and MR training results in equal workload. We could not reject our null hypothesis in total as well as in per category workload between MR and video by performing a two-sample two-tailed t-test using a significance level $\alpha = 0.05$. This, in HCI commonly chosen significance level, is important to minimize the uncertainty.  %We used the significance level $\alpha = 0.05$ to ensure that no more than 5\% of the time, the result is not better than random.

\definecolor{blue1}{RGB}{100,180,255}
\definecolor{red1}{RGB}{255, 60, 60}

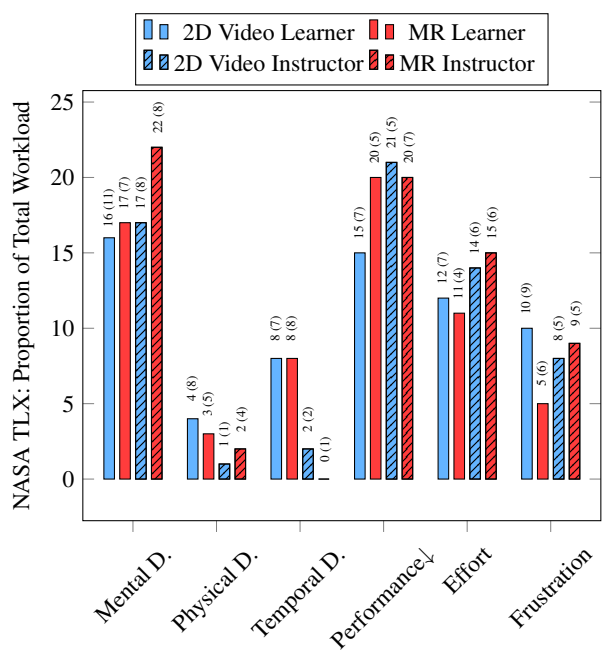
\begin{figure}%[!ht]
    \centering
    %\resizebox{\columnwidth}{!}{%
\begin{tikzpicture}  
\begin{axis}  
[  bar width=4, % NEW BIT
    ybar, % ybar command displays the graph in horizontal form, while the xbar command displays the graph in vertical form.  
    enlargelimits=0.12,% these limits are used to shrink or expand the graph. The lesser the limit, the higher the graph will expand or grow. The greater the limit, the more graph will shrink.   
    legend style={at={(0.5,1.18)}, % these are the measures of the bottom row, where -0.25 is the gap between the bottom row and the graph.   
      anchor=north, legend columns=2},      
      % here, north is the position of the bottom legend row. You can specify the east, west, or south direction to shift the location.   
    ylabel={NASA TLX: Proportion of Total Workload}, % there should be no line gap between the rows here. Otherwise, latex will show an error.  
    ylabel style={},
    yticklabel style={},
    ymax=23,
    symbolic x coords={Mental D.,Physical D., Temporal D., Performance$\downarrow$, Effort, Frustration},  
    xtick=data,  
    xtick pos=left,
    xticklabel style={rotate=45},
    %nodes near coords=\pgfmathsetmacro{\mystring}{{\mylst}[\coordindex]}\mystring,  
    nodes near coords align={horizontal}, 
    nodes near coords style={font=\small, rotate=90},
    width=\columnwidth,
    ]  
% Student 2D
\addplot[fill={blue1}] coordinates {(Mental D., 16) (Physical D., 4) (Temporal D., 8) (Performance$\downarrow$, 15) (Effort, 12) (Frustration, 10) }; % these are the measures of a particular bar graph. The tick marks of the y-axis will be adjusted automatically according to the data values entered in the coordinates.  
\node [above,xshift=-0.13cm,yshift=0.4cm, style={font=\tiny, rotate=90}] at (axis cs:  Mental D., 16) {16 (11)};
\node [above,xshift=-0.13cm,yshift=0.4cm, style={font=\tiny, rotate=90}] at (axis cs:  Physical D., 4) {4 (8)};
\node [above,xshift=-0.13cm,yshift=0.4cm, style={font=\tiny, rotate=90}] at (axis cs:  Temporal D., 8) {8 (7)};
\node [above,xshift=-0.13cm,yshift=0.4cm, style={font=\tiny, rotate=90}] at (axis cs:  Performance$\downarrow$, 15) {15 (7)};
\node [above,xshift=-0.13cm,yshift=0.4cm, style={font=\tiny, rotate=90}] at (axis cs:  Effort, 12) {12 (7)};
\node [above,xshift=-0.13cm,yshift=0.4cm, style={font=\tiny, rotate=90}] at (axis cs:  Frustration, 10) {10 (9)};

%Student MR
\addplot[fill={red1}] coordinates {(Mental D., 17) (Physical D., 3) (Temporal D., 8) (Performance$\downarrow$, 20) (Effort, 11) (Frustration, 5) }; 
\node [above,xshift=0.08cm,yshift=0.4cm, style={font=\tiny, rotate=90}] at (axis cs:  Mental D., 17) {17 (7)};
\node [above,xshift=0.08cm,yshift=0.4cm, style={font=\tiny, rotate=90}] at (axis cs:  Physical D., 3) {3 (5)};
\node [above,xshift=0.08cm,yshift=0.4cm, style={font=\tiny, rotate=90}] at (axis cs:  Temporal D., 8) {8 (8)};
\node [above,xshift=0.08cm,yshift=0.4cm, style={font=\tiny, rotate=90}] at (axis cs:  Performance$\downarrow$, 20) {20 (5)};
\node [above,xshift=0.08cm,yshift=0.4cm, style={font=\tiny, rotate=90}] at (axis cs:  Effort, 11) {11 (4)};
\node [above,xshift=0.08cm,yshift=0.4cm, style={font=\tiny, rotate=90}] at (axis cs:  Frustration, 5) {5 (6)};

% Teacher 2D
\addplot[postaction={pattern=north east lines}, fill={blue1}] coordinates {(Mental D., 17) (Physical D., 1) (Temporal D., 2) (Performance$\downarrow$, 21) (Effort, 14) (Frustration, 8)}; 

\node [above,xshift=0.30cm,yshift=0.4cm, style={font=\tiny, rotate=90}] at (axis cs:  Mental D., 17) {17 (8)};
\node [above,xshift=0.30cm,yshift=0.4cm, style={font=\tiny, rotate=90}] at (axis cs:  Physical D., 1) {1 (1)};
\node [above,xshift=0.30cm,yshift=0.4cm, style={font=\tiny, rotate=90}] at (axis cs:  Temporal D., 2) {2 (2)};
\node [above,xshift=0.30cm,yshift=0.4cm, style={font=\tiny, rotate=90}] at (axis cs:  Performance$\downarrow$, 21) {21 (5)};
\node [above,xshift=0.30cm,yshift=0.4cm, style={font=\tiny, rotate=90}] at (axis cs:  Effort, 14) {14 (6)};
\node [above,xshift=0.30cm,yshift=0.4cm, style={font=\tiny, rotate=90}] at (axis cs:  Frustration, 8) {8 (5)};

% Teacher MR
\addplot[ postaction={pattern=north east lines}, fill={red1}] coordinates {(Mental D., 22) (Physical D., 2) (Temporal D., 0) (Performance$\downarrow$, 20) (Effort, 15) (Frustration, 9)};
\node [above,xshift=0.53cm,yshift=0.4cm, style={font=\tiny, rotate=90}] at (axis cs:  Mental D., 22) {22 (8)};
\node [above,xshift=0.53cm,yshift=0.4cm, style={font=\tiny, rotate=90}] at (axis cs:  Physical D., 2) {2 (4)};
\node [above,xshift=0.53cm,yshift=0.4cm, style={font=\tiny, rotate=90}] at (axis cs:  Temporal D., 0) {0 (1)};
\node [above,xshift=0.53cm,yshift=0.4cm, style={font=\tiny, rotate=90}] at (axis cs:  Performance$\downarrow$, 20) {20 (7)};
\node [above,xshift=0.53cm,yshift=0.4cm, style={font=\tiny, rotate=90}] at (axis cs:  Effort, 15) {15 (6)};
\node [above,xshift=0.53cm,yshift=0.4cm, style={font=\tiny, rotate=90}] at (axis cs:  Frustration, 9) {9 (5)};

\legend{2D Video Learner, MR Learner, 2D Video Instructor, MR Instructor}  
  
\end{axis}  
 \end{tikzpicture}  
%       \caption{NASA TLX chart. We compare the mean weighted NASA TLX score for learner and instructor using video and MR teaching. We present the score on the y-axis and the cognitive load categories on the x-axis. We represent the video and MR results using blue and red bars, respectively. The instructor bars are filled with a line pattern.}
%       \label{fig:nasa-tlx}
%\end{figure}
       \caption{NASA TLX workload. We compare the weighted NASA TLX score (y-axis) per category (x-axis) for learner and instructor using video and MR teaching. We represent the video and MR results using blue and red bars, respectively. The instructor bars are filled with a line pattern. The standard deviation is shown in parenthesis.}
       \label{fig:nasa-tlx}
\end{figure}

\paragraph{Instructor Feedback} 
From interviews and observation of the instructor, we learned that each volumetric view, US view, and 2D view are essential during different parts of the procedure and for different purposes. The instructors liked the volumetric view because it gave them a spatial understanding of the scene and it allowed for visual communication using gestures and objects. However, smaller and translucent surfaces were sometimes not captured correctly in the volumetric view making it difficult for instructors to identify them. %The depth sensor has difficulty capturing very small and translucent objects due to the limitations of the technology. Thus, these objects don’t appear correctly on the volumetric view and the
To overcome this issue, the instructors used the video feed as a backup. We also observed that virtual objects in combination with gestural communication and the volumetric view can be used to effectively teach the correct usage of medical equipment. This turned out to be especially useful for teaching needle-probe coordination. The US feed together with the volumetric view gave the remote instructor a good spatial understanding of the needle-probe guidance of the local learner during an essential part of the procedure.   

\paragraph{Learner Feedback} 
The learners reported that the augmented instructor's hands and objects helped them learn how to use the medical equipment much faster than using verbal instruction only. However, the learners also reported that the teachers augmented hands sometimes were distracting because they were visible throughout the procedure and it was not clear if they are actively used by the instructor for communication. This suggests  minor modifications in the software.  Moreover, the learners highlighted the importance of the instructor's webcam view to see who they are communicating with. The opinions on the augmented US feed were mixed. Some learners liked it because it was positioned very close to the procedure area. Others preferred to use the physical US screen.  

\subsection{Discussion}
The fact that the subjective workload was not significantly higher is considered a positive result for the design of the system given that it was the first time that the learners used XR (AR, MR, VR) and the Hololens 2. Because video conferencing technology is used in everyday life it is expected to produce a lower extraneous workload. Although the instructors were familiar with the basic functionality of the system before they first used it, they did not have full training with the system before their first study session. Thus, we argue that our system is very intuitive to use and it only requires a five-minute hardware and user interface debrief before using it. 

We observed a learning effect of the instructor teaching through MR. The more often they taught, the more they utilized virtual objects and hand gestures. Hence, MR with experienced users might result in a lower workload and a better experience.  

Analyzing the individual workload categories, we argue that the trend toward higher frustration for the learner using video came from the fact that complex parts of the procedure were harder to understand using 2D visuals and needed multiple iterations of explaining. 
A reason for the high mental demand for the instructor can be that mixed reality has many options to teach and observe.   

\section{Conclusion}
We presented a mixed reality (MR) real-time communication system for assistance during ultrasound-guided central line placement. The system allows a remote expert to connect to a local operator to guide them in MR. The MR interaction allows for vocal and gestural communication. The communication is based on volumetric capture through RGBD cameras that allow for intuitive visual guidance. 
We proposed an algorithm that focuses on lightweight, real-time communication and rendering of volumetric capture. 

We evaluated the proposed system in a user study in which we compare MR against video communication for ultrasound-guided CVC placement training. We found that MR provides a viable alternative compared to video during the CVC procedure training. We showed how the different elements of the system can be used effectively during procedural training.  %Our MR system is easy to setup, intuitive to use and simplifies visual communication. %In future work, we plan to evaluate performance in medical procedures comparing MR against other technologies.

%Volumetric communication in medical training context is a new approach, requiring further research. While in this paper, we presented an elicitation study and prototype system, more work is needed to understand whether the volumetric capture technology application to the medical procedure training is feasible, and whether it results in the same learning outcomes as in person training.

While we focus on a single medical procedure, the issue of training and assistance by an expert who is not on-site is present across multiple disciplines and many domains which depend on operator cognitive and manual procedural skills \cite{hadar2011hybrid, rebol2021remote}. When mastery-level skill must be brought to a remote location during natural disasters, epidemics, equipment breakdowns, etc., our system could be sent to the remote location or travel with providers who may be in need of support.

\acknowledgments{
The work is supported by National Science Foundation grant no. 2026505 and 2026568. 
The authors wish to thank Erin Horan, Safinaz Alshiakh, Yasser Ajabnoor, Ahmed Allabban, Becky Lake, Scott Schechtman, Rahil Ashraf, and Carine Cristina Goncalves Galvao for their help.
Moreover, the authors would like to thank medical students and residents for participating in the experiment.}
% Rohan, Conor in Hicss paper.

%\bibliographystyle{abbrv}
\bibliographystyle{abbrv-doi}

\bibliography{paper}

\renewcommand{\thesubsection}{\Alph{subsection}.}

\section*{Appendix}
\subsection{3D grid lookup}

We first read the depth value and then the color value for every point $P$ on the grid $\pmb{V}$.
The depth allows us to compute the 3D position of a point in the scene. We take every linearly distributed point $P=(X,Y,Z): X \in [-0.5,+0.5], Y \in [-0.5,+0.5], Z = 1$ in real-world coordinates. We define the depth camera coordinates to be the same as the real-world coordinates. Thus, for every real world point $P=(X,Y,Z)$ we get the corresponding depth camera coordinate point $P_d=(x,y,z) = (X,Y,Z)$. Then, we distort this point according to the lens distortion coefficients provided by the Kinect SDK. $k_1, k_2, k_3, k_4, k_5, $ and $k_6$ are radial distortion coefficients and $p_1$, and $p_2$ are tangential distortion coefficients. We compute the undistorted depth image plane coordinates $x''$ and $y''$ using the equations: 
\begin{equation}\label{eqn:distort}
\begin{aligned} 
x' &= x/z, \\ 
y' &= y/z, \\ 
r^2 &= x'^2 + y'^2, \\  
x'' &= x' \frac{1+k_1r^2 + k_2r^4 + k_3r^6}{1 + k_4r^2 + k_5r^4 + k_6r^6} +2p_1x'y' + p_2(r^2, 2x'^2),\\
y'' &= y' \frac{1+k_1r^2 + k_2r^4 + k_3r^6}{1 + k_4r^2 + k_5r^4 + k_6r^6} +p_1(r^2 + 2y'^2) + 2p_2x'y' .
\end{aligned}
\end{equation}
Using the depth camera intrinsic parameters consisting of the principal point $(cx,cy)$ and the focal lengths in pixel units $fx, fy$, we compute the depth image pixels $(u,v)$ as follows:
\begin{align} \label{eqn:intrinsic}
    u &= f_x \cdot x'' + c_x, \\
    v &= f_y \cdot y'' + c_y. \nonumber
\end{align}
Finally, we retrieve the depth value $d \in [0,2^{16}]$ in millimeters for real-world point $P$ from the depth image $\mathcal{D}(u,v) = d$.

Once we found the depth value for the real-point $P$, we compute the color value similarly. First, we compute the color lens distorted real-world coordinates $(x''_c,y''_c)$ using \Cref{eqn:distort}. Then, we  convert the resulting point $(x''_c,y''_c, Z)$ from the real-world into the color camera coordinate system. We use the extrinsic parameters of the color camera $\pmb{R}_c \in \mathbb{R}^{3 \times 3}$ and $t_c \in \mathbb{R}^3$:
\begin{equation}
    P_c = (x_c,y_c,z_c)^T = \pmb{R}_c P+ t_c .
\end{equation}
Once we switched to the color camera coordinate system, we project $P_c$ onto the color image plane:
\begin{align} \label{eqn:color-project}
    x_{cp} &=  x_c / z_c, \\
    y_{cp} &=  y_c / z_c . \nonumber
\end{align}
We get the color image coordinates after applying the color camera intrinsic parameters similar to \Cref{eqn:intrinsic}.
Finally, we lookup the color value $c = (r,g,b) \in \mathbb{N}^3$ for real-world point $P$ from the color image $\mathcal{C}(u_c,v_c) = (r,g,b)$.
As a result of the color and depth image pixel lookup, we get the 3D position of each vertex in our grid $v = (X, Y, 1)\cdot d \cdot 10^{-3}$ in meters relative to the depth sensor and its color value $c$. 

% As a result of the color and depth image pixel lookup, we get the 3D position of each vertex on our grid in meters relative to the depth sensor and its color value. 
% We connect neighboring vertices on our grid to create a mesh. However, we only connect vertices less than 0.1 m away from each other. We linearly interpolate the color information from the image onto the mesh. 

\subsection{Vertex manipulation}
\begin{figure}[h!]
    \centering
    \includegraphics[width=\columnwidth]{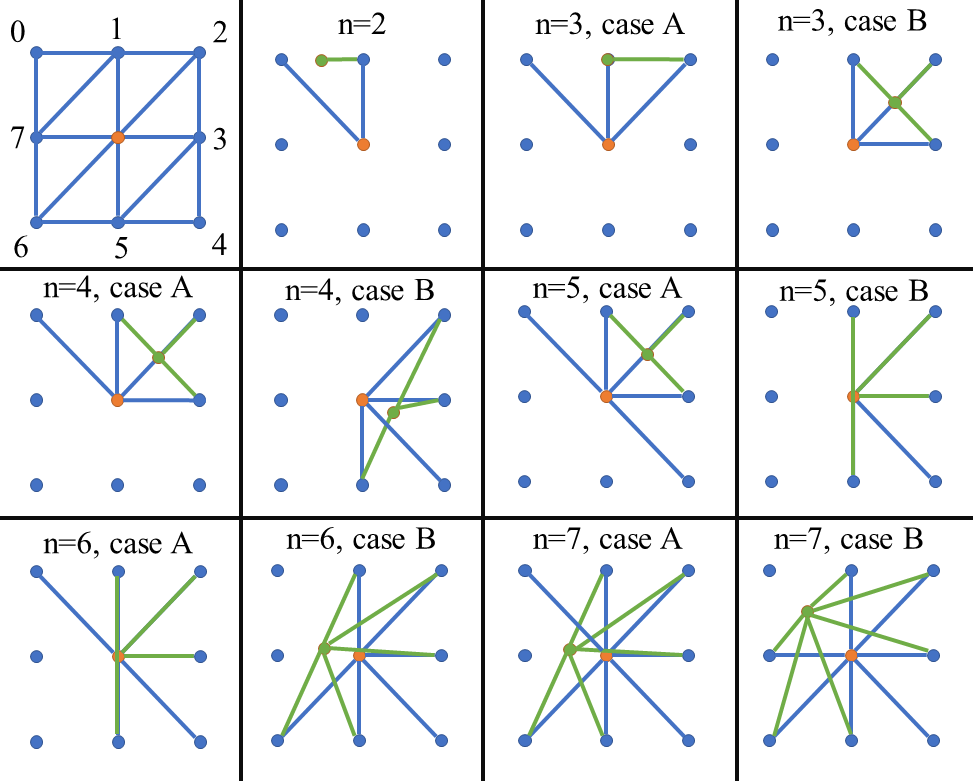}
    \caption{Vertex manipulation. We consider each central vertex (orange dot) and 8 neighboring vertices (blue dot) numbered clock-wise starting at the top left. The first square (top right) shows the triangulation topology to create a mesh. The other squares show how we modify the central vertex position to refine the edges of the mesh taken into account the grid topology. Depending on how many neighbors $n$ a vertex has, a new vertex position (green dot) is assigned. We show two example cases A and B for the per neighbor count. Blue lines indicate original neighbor presence and green lines indicate the final mesh connections from the central vertex after it was moved. Note that this is a 2D representation of the 3D grid $V$.}
    \label{fig:nh}
\end{figure}

We show our grid topology and examples of how we modify the vertices depending on the number of neighbors in \Cref{fig:nh}.

\end{document}